\documentclass[10pt,conference]{IEEEtran}
\IEEEoverridecommandlockouts
\usepackage{cite}
\usepackage{amsmath,amssymb,amsfonts}
\usepackage{algorithmic,authblk}
\usepackage{graphicx,booktabs,makecell,enumitem}
\usepackage{textcomp}
\usepackage{xcolor}
\usepackage{array}
\usepackage{enumitem,cleveref}

\newcommand{\xt}[1]{\texttt{#1}}
\def\BibTeX{{\rm B\kern-.05em{\sc i\kern-.025em b}\kern-.08em
    T\kern-.1667em\lower.7ex\hbox{E}\kern-.125emX}}
\usepackage{comment}

\begin{document}
\author[*]{Charalampos Kleitsikas}
\author[$\dagger$]{Nikolaos Korfiatis}
\author[*]{Stefanos Leonardos}
\author[*]{Carmine Ventre}

\affil[*]{Department of Informatics, King's College London, United Kingdom}
\affil[$\dagger$]{Nottingham University Business School, University of Nottingham, United Kingdom}

\affil[*]{\{charalampos.kleitsikas, stefanos.leonardos, carmine.ventre\}@kcl.ac.uk}
\affil[$\dagger$]{nikolaos.korfiatis@nottingham.ac.uk}
\title{Bitcoin's Edge: Embedded Sentiment in\\ Blockchain Transactional Data}

\maketitle

\begin{abstract} 
Cryptocurrency blockchains, beyond their primary role as distributed payment systems, are increasingly used to 
store and share arbitrary content, such as text messages and files. Although often non-financial, this hidden content can 
impact price movements by conveying private information, shaping sentiment, and influencing public opinion. However, current analyses of such data are limited in scope and scalability, primarily relying on manual classification or hand-crafted heuristics. In this work, we address these limitations by employing Natural Language Processing techniques to analyze, detect patterns, and extract public sentiment encoded within blockchain transactional data. Using a variety of Machine Learning techniques, we showcase for the first time the predictive power of blockchain-embedded sentiment in forecasting cryptocurrency price movements on the Bitcoin and Ethereum blockchains. Our findings shed light on a previously underexplored source of freely available, transparent, and immutable data and introduce blockchain sentiment analysis as a novel and robust framework for enhancing financial predictions in cryptocurrency markets. Incidentally, we discover an asymmetry between cryptocurrencies; Bitcoin has an informational advantage over Ethereum in that the sentiment embedded into transactional data is sufficient to predict its price movement.  

\end{abstract}


\section{Introduction}
First introduced by the Bitcoin software \cite{nakamoto2009}, cryptocurrency blockchains are electronic payment systems that eliminate the need for centralized authorities, such as banks, to validate transactions. Due to their core principles --decentralization, the removal of intermediaries, transparency, and transaction security among others-- blockchains have experienced exponential growth in popularity over the past two decades. However, beyond being simple financial transaction records, blockchains have evolved, due to their key feature of immutability, to serve as repositories for arbitrary content, including text messages, images, and, in some cases, controversial material \cite{matzutt2016, matzutt2018}. This content is encoded and embedded into the blockchain either by users or through specialized services, using low-level data insertion methods (see, e.g., \Cref{tab1}).\par 
The first instance of arbitrary content in Bitcoin was hardcoded into the blockchain by Nakamoto himself in the Genesis block, the very first mined block. Encoded within the coinbase transaction (a non-standard transaction that rewards the miner with newly minted coins) is the string: “\xt{The Times 03/Jan/2009 Chancellor on brink of second bailout for banks.}” Miners often encode messages in the input scripts field of the coinbase transaction, commonly for purposes such as advertising mining pools or unofficial feature voting. Non-miner users can also embed content in various ways, including through \xt{Pay-to-Pubkey} (P2PK) and \xt{Pay-to-Pubkey-Hash} (P2PKH) transactions, Nulldata transactions, the \xt{OP\_RETURN} field, or by creating ASCII addresses without a known private key to burn coins \cite{Shirriff2014}. In addition, numerous web platforms have been created to offer the opportunity to embed non-transactional data, such as wishes, proof of existence, messages, in blockchains, e.g., \xt{Cryptograffiti}, \xt{Proof of Existence} and \xt{Agora} \cite{sward2018}.\par
However, despite being permanent and visible to all participants, the context of these embedded messages has remained largely underexplored. Although recent studies have developed sophisticated decoding and text extraction tools, see, e.g., \cite{gregoriadis2022,matzutt2018}, existing analyses of the extracted content primarily rely on hand-crafted heuristics, such as frequency analysis or qualitative human inspection, leaving a significant portion of messages without proper contextual evaluation \cite{bartoletti2017, bartoletti2019, scheid2023, rodwald2021}. Such methods, while important in categorizing the data and assessing its growth over time, are not suitable for applications that require automatic context inference, e.g. fraudulent and illicit content detection, phishing and scam alerts, sentiment-driven trading etc., nor for comprehensive offline research due to their non-exhaustive nature.
Our first contribution in this paper addresses these limitations.
\begin{enumerate}[label=\arabic*., wide, labelindent=0pt]
    \item We use Natural Language Processing  (NLP) techniques to analyze the arbitrary text that is embedded in the Bitcoin and Ethereum transactional data. We apply topic modeling, specifically, \textit{Latent Dirichlet Allocation} (\textit{LDA}) \cite{blei2006}, \cite{tharwat} and \textit{BERTopic} \cite{grootendorst2022}, to automatically discover hidden semantic patterns and identify topics that exist within the blockchains text corpuses (see \Cref{nlp_pipeline}). While both the broad-themed LDA and the more context-rich BERTopic discover clusters of foundational or programming topics in both cryptocurrencies, e.g., \xt{peer}, \xt{electronic}, \xt{function}, \xt{test}, they also uncover direct financial or privacy-related and transactional themes, e.g., \xt{private}, \xt{buy}, \xt{today}, \xt{signal}, often also related to altcoins, e.g., \xt{doge}, \xt{luna}, hidden, particularly, in the Bitcoin blockchain (\Cref{tab2}).
\end{enumerate}

These findings suggest that users intentionally, and often, systematically use embedded blockchain messages to (anonymously) share insights, shape sentiment or provide hidden trading signals regarding cryptocurrency prices movements. This naturally prompts the compelling question of whether such non-financial content, which may, nevertheless, carry implicit financial context can have predictive power over price dynamics. While extensive research has analysed the influence of social media, news and general public sentiment found across the internet on predicting crypto prices \cite{raju2020, georgoula2015, sattarov2020, loginova2024, kilimci2020, kumari2023, coulter2022, Bhatt2023, passalis2022, gurrib2022}, to the best of our knowledge, no study has explored the potential impact of sentiment hidden in arbitrary embedded blockchain messages. In this paper, we aim to bridge this gap, and make the following two key contributions in this context. 
\begin{enumerate}[resume, label=\arabic*., wide, labelindent=0pt]
\item We apply Sentiment Analysis to capture the attitudes, opinions, and emotions that are hidden in the embedded blockchain messages. We use rule- and lexicon-based methods, such as VADER \cite{hutto2014} and TextBlob \cite{loria2018textblob}, as well as transformer-based models (e.g., BERT \cite{devlin2019}), fine-tuned and pretrained on extensive collections of crypto-related financial data. Motivated by our topic modeling analysis, which highlights a wide range of topics discussed with varying intensities within blockchains (see Figure \ref{fig1}), we investigate whether domain-agnostic approaches (VADER and TextBlob) provide more accurate insights than specialized models (CryptoBERT) for analyzing on-chain sentiment. Our findings indicate that blockchain topics and sentiment are not only predominantly Bitcoin-focused compared to Ethereum, but also that rule-based approaches outperform transformer-based models in extracting sentiment and quantifying its influence on price movements for both cryptocurrencies (see Section \ref{sec:discussion}). 
\item We, then, proceed to quantify the predictive power of the derived text sentiment on cryptocurrency price movements by deploying a wide range of Machine Learning (ML) model Classifiers, including \textit{Random Forests (RFC), XGBoost (XGBC), K-Nearest Neighbors (KNNC), AdaBoost (ABC)} and \textit{Extra Trees (ETC)} among others. For Bitcoin, when sentiment is solely used for 1-day ahead predictions, prediction accuracy surpasses all of our baseline algorithms and achieves an improvement of $6.25\%$ compared to random guess. When combined with price data the improvement increases to $10.53\%$. By contrast, Ethereum's predictions fail to beat the baselines when ML models are provided only with sentiment features, showcasing that blockchain hidden messages mainly influence Bitcoin prices (\Cref{tab:metrics_side_by_side}). 
\end{enumerate}
With these findings, our work extends existing literature, e.g., \cite{gurrib2022, loginova2024, passalis2022}, by demonstrating comparable improvements in understanding the impact of sentiment on cryptocurrency price movements. Importantly, however, this is achieved using a previously unexplored data source, for sentiment analysis. Unlike platforms such as Twitter or Reddit, this source is freely accessible, uncensored, and transparent, laying the foundation for a more comprehensive understanding of the volatile cryptocurrency ecosystem.



The remaining of this work is organized as follows. Section \ref{sec:literature} provides an extensive literature review regarding arbitrary blockchain content detection and analysis, and the existing types of studies that use sentiment analysis for predicting cryptocurrencies’ prices. 
In section \ref{sec:method} a detailed and thorough approach of our methodology is described. The results of our experiments and their evaluation is in section \ref{sec:res}. Lastly, section \ref{sec:conclusion} has the conclusions of our work.

\section{Literature Review} \label{sec:literature}
To the best of our knowledge, no systematic framework for classifying blockchain text messages based on their context has yet been developed. More closely related to our analysis are the works of \cite{matzutt2016,matzutt2018}, \cite{bartoletti2017} and \cite{matzutt2022}. \cite{matzutt2016} studied the evolution of transactions with arbitrary data for different storage methods and found encrypted files, images, and malicious content such as leaked private keys and illegal pornography links by manually searching the dataset. \cite{matzutt2018} created detectors for the different insertion methods and focused their analysis on readable files, again manually examined their derived dataset. The files mainly contained copyright and privacy violations, malware, and politically sensitive and illegal content. \cite{matzutt2022} proposed a framework for quick and clear moderation of content on the blockchain. Finally, \cite{bartoletti2017} explored the use of the OP\_RETURN field in Bitcoin transactions and investigated how this feature has been utilized, categorizing the data and assessing its growth over time. In general, the above works mainly focus on the development of content detectors and on finding ways to mitigate arbitrary data insertions. Thus, in contrast to our work, they don't provide quantitative analysis of the content's context per se.\par
\cite{gregoriadis2022} developed a cloud-based approach for discovering and classifying content, mainly in terms of insertion method, on the bitcoin and ethereum blockchains. They also investigated the most frequently embedded texts found on blockchains and made qualitative and quantitative observations about these as well as for URLs, file types and images. Similarly to \cite{matzutt2016, matzutt2018}, they didn't evaluate in terms of context every arbitrary content they decoded on Bitcoin or Ethereum but manually a sample of them. On the same note, \cite{rodwald2021} manually analysed types of messages hidden in identified Bitcoin and Ethereum addresses. \par
The use of on-chain data to predict cryptocurrency price movements is a common theme in existing literature. However, most studies either mainly focus on their aggregated properties \cite{politis2021, saad2018} (e.g. the total number of transactions per time step) or 
utilize variants of the blockchains' transaction network graphs, which use with Machine Learning (ML) models to predict crypto prices, see, e.g., \cite{akcora2018, li2020blockchaintransactiongraphbased, kleitsikas}  However, such studies mine on-chain data that are unrelated to the arbitrary embedded text within blockchains. \par
Finally, there is also a rich literature regarding price prediction and financial analysis for cryptocurrencies, which considers sentiment analysis based on social media, news websites and online forum platforms \cite{raju2020, georgoula2015, sattarov2020, Kok22, loginova2024, kilimci2020, kumari2023, coulter2022, Bhatt2023}. Most of these works primarily concentrate and process tens of thousands or even millions of text messages from online websites, such Twitter (officially known as "X" since July 2023), Reddit, Telegram, Bitcointalk, Factset, and Cryptocompare, transforming them into features which are then fed to predictive models (mainly ML models) for price forecasting. However, with only few exceptions, the reported data sources either require professional or paid subscriptions to their APIs or have quotas to their free usage, see e.g., \cite{passalis2022, gurrib2022}. In contrast to these data sources, permissionless blockchains offer a unique communication channel. Unlike platforms where pricing and policies frequently change, blockchains ensure transparency, immutability as well as free and permanent access to the sentiment, content or messages that are embedded in their transactions. Our work develops the first systematic analysis of this previously unexplored direction, providing a stable and sustainable approach in blockchain sentiment extraction.

\section{METHODOLOGY}\label{sec:method}

\begin{figure}[tbp]
\centering
\begin{minipage}[b]{\columnwidth}
    \centering
    \includegraphics[width=1\linewidth]{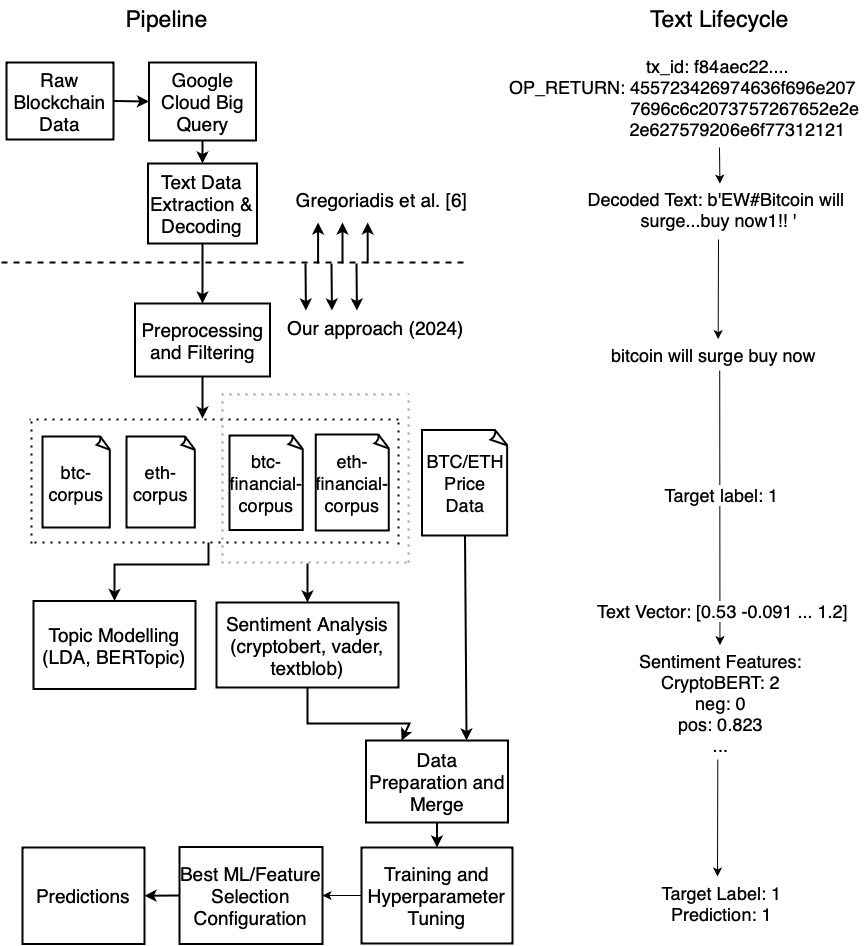}
\end{minipage}
\caption{Our proposed NLP pipeline.}
\label{nlp_pipeline}
\end{figure}

The overall high-level approach of our proposed pipeline, as well as an example of a text's transformation throughout it, can be seen in Fig.\ref{nlp_pipeline}. For the text extraction and decoding we use the tool developed by \cite{gregoriadis2022}. After obtaining the text corpuses, due to their noisy nature, we apply extensive preprocessing and filtering steps to obtain different versions of the texts. We provide two different types of results. First, by using Topic Modeling we extract the most salient topics alongside with their accompanying keywords for every corpus. That way, we have an automatic thematic classification of the corpuses that avoids the time-consuming process of going manually through texts. Second, we make daily predictions for the price movements of Bitcoin and Ethereum. To do that, we extract sentiment features from the created sub-corpuses that focus on financial context and after merging them with the price data of these two cryptocurrencies, we train ML models under different feature combinations. The best ones are used to obtain the predictions in our test set and evaluate our approach. Our code and data are open-sourced and can be found in this GitHub repository\footnote{https://github.com/kleitsikas/embedded\_sentiment\_blockchain}.

\subsection{Data Collection}

There are numerous approaches to embed encoded arbitrary content into the blockchains, e.g., in Bitcoin this can be done via the OP\_RETURN output scripts, the coinbase input scripts for the miners, non-standard transactions, or pay-to-script-hash scripts among others. To decode and extract this content, appropriate detectors must be developed, which will filter the raw transactional data and keep the transactions that contain text content. To that end, we use the detector toolkit that was developed by \cite{gregoriadis2022} and consider the data inserted through general, low-level insertion techniques. \cite{gregoriadis2022} developed their methodology for the raw transaction datasets of both Bitcoin and Ethereum that are uploaded to the Google BigQuery platform and are regularly kept up to date. By setting up their toolkit and enabling the BigQuery API, we obtain our initial datasets that contain the detectable texts inside transactional metadata for both Bitcoin and Ethereum, ranging from 03-Jan-2009 to 06-Nov-2023 for Bitcoin and from 07-Aug-2015 to 06-Nov-2023 for Ethereum. We also download the daily price data of Bitcoin and Ethereum from Bloomberg's terminal, which range from 11-Nov-2017 to 10-Jun-2023. For our first task, which involves the contextual classification of the embedded content in the blockchain, we use the entirety of the initial text datasets. For the second task which involves determining the public sentiment embedded in the blockchain and the quantification of its predictive power in the cryptocurrency market, the texts date range is limited to that of our available price data.

\subsection{Data Inspection}

The initial dataset contains many transaction related fields such as \textit{'hash'}, which is the transaction’s unique ID, \textit{‘data’}, which contains the embedded text, ‘value’, which contains the value of the transaction that has been made, \textit{‘block\_timestamp’}, indicating the time that the block which contains that transaction was created etc. The fields that we will be focusing on are the \textit{‘data’} and \textit{‘block\_timestamp’}, as the rest are not useful for our analysis. 

We can easily observe from our overall inspection of the datasets, that the \textit{‘data’} field has highly unstructured data with a lot of noise. A lot of symbolic characters are evident, long strings without white-spaces containing alphanumeric characters ranging from base64 encoded formats to transaction identifiers, JSONs etc. This indicates the need for an extensive clean-up and filtering of our texts to make them suitable for our NLP models. The next section is dedicated to that purpose.

Indicatively, Table \ref{tab1} contains some messages found both in the Bitcoin and Ethereum blockchains, containing both unprocessed strings and text containing trading signals. Our core assumption is that such content that would potentially impact the prices of the cryptocurrencies exists in the blockchains (e.g., buy and sell signals).


\begin{table}[htbp]
\caption{Sample Text Data}
\centering
\begin{tabular}{@{}p{6cm}p{2.4cm}@{}}
\toprule
\textbf{Text} & \textbf{Block Timestamp} \\
\midrule
\makecell[tl]{SWAP:ETH.ETH:0xff6763c12c4cc54b2fd4115\\
690d7792ecb78bd55:242384111} & 2021-05-24 10:59:18 \\
\midrule
\makecell[tl]{"T.""ver"":2,""cmd"":""CU"",\\
""tx\_list"":[""636969.1529""]"} & 2020-07-02 00:27:12 \\
\midrule
Saturday, March/05/2016 - Signal for today is 100 \% BTC / 0 \% USD & 2016-03-05 19:32:47 \\
\midrule
Signal for today is 0 \% BTC / 100 \% USD & 2015-08-09 19:41:18 \\
\midrule
EW ETH pumping, BTC dumping. Should I buy ETH, BTC or hold fiat? & 2016-01-23 12:17:42 \\
\midrule
hodl on comrade! & 2020-03-04 02:23:19 \\
\midrule
Sell FTSE Mib ITALY Start a new short trade - Net Position Short 29.09 Cash 38.18 & 2016-05-02 08:41:54 \\
\midrule
Happy Birthday.... He sold? Pump It. & 2022-01-05 06:14:37 \\
\bottomrule
\end{tabular}
\label{tab1}
\end{table}

Signals like the ones of the third and fourth rows appear around 390 times in our dataset. Other financial signals like the rest shown in the table seem to not follow a systematic format. These texts are a prime example of what we are looking for to capture sentiment and explore how this affects the price formation of cryptocurrency prices. Of course, sentiment of other types of messages on the blockchain, such as news for new regulatory measures or wars, problems in transaction procedures, lost coins etc. also impact prices and should be taken into consideration. \par

From our price datasets, we will use the \textit{‘Date’}, and \textit{‘Close’} fields. The latter one will be used to produce our target labels (0 if the 1 day ahead close price is lower than the current close price of the current row and 1 if it is higher).

\subsection{Text Cleanup and Preprocessing} \label{4c}

Our corpuses are highly populated with data noise. This could have a significant negative impact on the performance and effectiveness of our NLP models used for sentiment and thematic analysis. To avoid this, we transform our text data into a clean and consistent format, suitable to be processed by our models. To that end, for both our text corpuses (we will use the terms \textit{btc-corpus} and \textit{eth-corpus} respectively throughout the rest of the paper), we perform a series of filtering and preprocessing steps, which can be summarized as follows:
\begin{enumerate}[leftmargin=*, label=\arabic*.]
    \item Remove artifacts (such as \texttt{b}' ' prefixes) caused by the original byte string format of the text data.
    \item Use regular expressions (generally, regex operations work by defining search patterns and rules using a combination of characters and metacharacters) to remove certain characters or substrings in a text or filter out texts completely if they satisfy certain properties.
    \item Filter out base64 encoded patterns, which are the strings that end with the double character \textit{‘==’}.
    \item Remove links in the text as they do not contribute to the overall semantics of the corpuses.
    \item Remove symbols (special characters), as they offer little value in the understanding of the meaning of the text.
    \item Discard numbers that are between letters.
    \item Omit texts that have more than 25 consecutive characters without white spaces, as they almost exclusively contain noise of random sequences of alphanumeric characters. 
    \item Omit texts with length smaller than 2 characters or containing only numbers as they lack semantic context.
\end{enumerate}

After these steps, we standardize all strings by lowercasing them in order to have consistent representations for our models. Further, given that the pre-trained models we use for our NLP tasks will not capture and represent adequately in vectors abbreviations regarding the financial market of cryptocurrencies, we replace them with their full phrases, e.g., \xt{fomo}: fear of missing out, \xt{btd}: buy the dip, \xt{ico}: initial coin offering etc. Moreover, we alter crypto slang so as not to lose content, e.g., \xt{hodl}: hold, \xt{Memecoin}: meme coin etc.

To further proceed with our preprocessing, we use the \xt{spacy.io} library and its pipeline for the English language to do word tokenization. With the help of word tokenization for each text, we remove words that are OOV (out of vocabulary), meaning that they don’t belong to the English corpus that \xt{spacy} has been trained on. After that, our texts are ready to be fed to the pre-trained models for Topic Modeling, as we will see in Section \ref{5a}. For the goal of our second task, further steps are required. As mentioned, we want to use the derived sentiment of texts to predict the price movement of Bitcoin and Ethereum. Thus, content that is completely irrelevant to financial context, e.g., tributes, advertisements, spam messages etc., only adds noise to our data and should be removed. To that end, we construct a hand-crafted lexicon containing around 800 words related specifically to crypto-markets, trading and general financial terms. After the word tokenization, we then remove every text that doesn’t contain at least one of the words that belong to the lexicon and create two extra datasets; \textit{btc-financial-corpus} and \textit{eth-financial-corpus}. Hence, the sizes of our datasets change as follows: {\boldmath$\cdot$} \textit{(unfiltered) btc-corpus}: 1,112,421 rows, \textit{btc-corpus}: 82,577 rows, \textit{btc-financial-corpus}: 7,034 rows, {\boldmath$\cdot$} \textit{(unfiltered) etc-corpus}: 4,425,192 rows, \textit{eth-corpus}: 1,800,982 rows, \textit{eth-financial-corpus}: 102,227 rows.

\subsection{Data Preparation}

For the sentiment analysis part, the corpuses and price datasets must be aligned and transformed appropriately. First, we produce our target labels by assigning 0 if the 1-day ahead close price is lower than the current close price of the current row or 1 if it is higher. As we showed in subsection \ref{4c}, \textit{btc-financial-corpus} and \textit{eth-financial-corpus} are just $0.63\%$ and $2.3\%$ of the size of the initial datasets accordingly. 
Therefore, to have enough texts to base our price predictions, we make the design decision to merge the text data of both blockchains’ financial corpuses into a combined corpus and aggregate for each day the unique (to avoid noise via repetition) texts found in each blockchain in that timespan. These aggregated texts will be used to derive the sentiment of that specific day, which, in turn, will be used to make the 1-day ahead predictions for Bitcoin and Ethereum price movements respectively, by serving as input for the deployed ML models. 

\subsection{Sentiment Analysis}

Many NLP methods have been developed to extract and appropriately quantify sentiment from documents. In this work, we use two widely accepted approaches. Rule-based sentiment analysis and deep learning. Regarding the former, we use VADER \cite{hutto2014} and TextBlob \cite{loria2018textblob} and regarding the latter, we use two variations of BERT \cite{devlin2019}. VADER uses a lexicon of ranked sentiment-related words, where each word contributes to the overall score of the text. BERT, on the other hand, takes the entire sentence as input and processes it through multiple layers of transformer networks to generate contextually rich embeddings. These embeddings capture sentiment by considering the complexity of words within their context. However, the arbitrary content within blockchains requires the extraction of very domain-specific sentiments, as the language that is used in it, contains a lot of slang and crypto-specific terminology. For example, “\xt{pump}” and “\xt{dump}” words hold very specific financial sentiments that can not be captured with general-purpose pre-trained models. \par
To account for this, we employ two kinds of fine-tuned pre-trained domain-specific BERT variants: 
{\boldmath$\cdot$} ElKulako/cryptobert’s model, which is trained on a corpus of over 3.2 million unique cryptocurrency-related posts on social media, and {\boldmath$\cdot$} kk08/CryptoBERT’s model, which is a fine-tuned version of ProsusAI/finbert on custom crypto market sentiment dataset. Further, using TextBlob and VADER, we also compute the following features: \textit{subjectivity, polarity, compound, neg, neu, and pos}. Thus, for each days’ text of our final dataset, we compute the features shown in Table \ref{tab_features}. 

\begin{table}[tbp]
\centering
\caption{Sentiment Features}
\setlength{\tabcolsep}{3pt} 
\renewcommand{\arraystretch}{1.0} 
\begin{tabular}{@{}p{0.2\columnwidth}@{\hspace{2pt}}p{0.2\columnwidth}@{\hspace{10pt}}p{0.55\columnwidth}@{}}
\toprule
\textbf{Feature} & \textbf{Values} & \textbf{Description} \\
\midrule
\textit{cryptobert\newline\_sentiment} & 0 (bearish), \,\,\,\,\,\,\,\, 1 (neutral),  \,\,\,\,\,\,\,\,\,  2 (bullish) & Output of ElKulako/cryptobert’s model, labeling data with sentiment categories. \\
\midrule
\textit{cryptobert\newline\_sentiment\_2} & 0 (negative), \,\,\,\,\,\, 1 (positive) & Output of kk08/CryptoBERT’s model, labeling texts as negative or positive. \\
\midrule
\textit{subjectivity} & 0 (objective) to 1 (subjective) & TextBlob's quantified measure of personal opinion versus factual information. \\
\midrule
\textit{polarity} & \hspace{0.5pt}--1 (negative) to +1 (positive) & TextBlob's sentiment score indicating positive, neutral, or negative sentiment. \\
\midrule
\textit{compound} & \hspace{0.5pt}--1 (negative) to +1 (positive) & Vader's overall sentiment score for the text. \\
\midrule
\textit{neg} & 0 to 1 & Vader's proportion of text falling into the negative sentiment category. \\
\midrule
\textit{neu} & 0 to 1 & Vader's proportion of text falling into the neutral sentiment category. \\
\midrule
\textit{pos} & 0 to 1 & Vader's proportion of text falling into the positive sentiment category. \\
\bottomrule
\end{tabular}
\label{tab_features}
\end{table}

\begin{figure}[tbp]
\centering
\begin{minipage}[b]{\columnwidth} 
    \centering
    \includegraphics[width=0.85\linewidth]{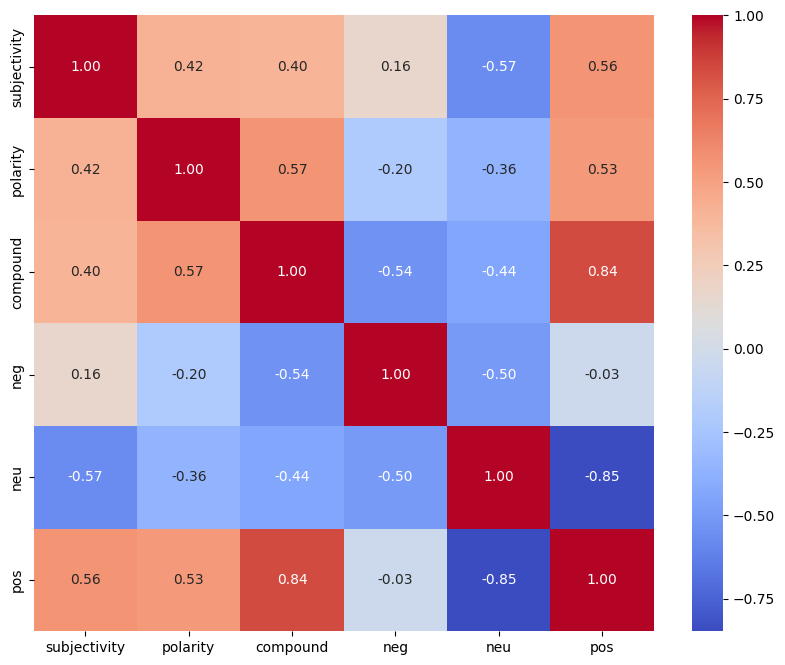}
\end{minipage}
\caption{Pearson Correlations.}
\label{fig2}\vspace*{-0.5cm}
\end{figure}

Given that the derived features are already normalized, we proceed with plotting the heatmap of Pearson's correlations of the continuous features as shown in Figure \ref{fig2}. We can see that the most highly linearly correlated features are \textit{compound} with \textit{pos}, having a correlation coefficient of 0.84. Also \textit{pos} with \textit{neu} are negatively highly correlated with a coefficient of -0.85. Thus, given that this strong linear correlation will not contribute to better results when these pairs are fed to our ML models, we remove the \textit{pos} feature.

\subsection{Machine Learning Pipeline} \label{MLsubsec}

To preserve the chronological order of our dataset, we split it to $85\%$ train set, i.e. from 11-Nov-2017 to 09-Aug-2022, and $15\%$ test set, i.e., from 10-Aug-2022 to 10-Jun-2023. As already mentioned, we use the aggregated sentiment of each day from both blockchains to predict the price movement of Bitcoin and Ethereum for the next day. 
The evaluation will be based on the results in the test set. For the train set, preserving again its chronological order, we perform a 5-fold timeseries cross-validation to allow ML models to train without overfitting their parameters on a validation set from a fixed timerange. To further avoid overfitting, we make sure that both our training and test sets are balanced in terms of class representation and, thus, of market conditions. For Bitcoin’s training set, we have $47\%$ of 0s and $53\%$ of 1s, and for its test set, we have $47.7\%$ of 0s and $52.3\%$ of 1s. Similarly, for Ethereum we have $48.6\%$ of 0s and $51.4\%$ of 1s for its training set, and $47\%$ of 0s and $53\%$ of 1s for its test set.\par

We use a wide range of ML classifiers, namely \textit{Random Forest (RFC), XGBoost (XGBC), K-Nearest Neighbors (KNNC), Logistic Regression (LRC), Support Vector (SVC), Gradient Boosting (GBC), AdaBoost (ABC)} and \textit{Extra Trees (ETC)}, from the scikit library. The chosen metric of optimization is accuracy and for hyperparameter tuning we use \textit{GridSearch}, for which all possible combinations of hyperparameters are explored with values given in Table \ref{tab3}. For the models that allow it, we also perform a \textit{Recursive Feature Elimination (RFE)} process \cite{guyon2002} to choose the sentiments that mostly influence the prediction accuracy in the validation sets while removing the ones that do not improve our results. Ultimately, these will be the features that will be used in our model to obtain our final classification results in the test set. The range of these are also shown in Table \ref{tab3}.

\begin{table}[tbp] 
\centering
\caption{Hyperparameters for GridSearch}
\setlength{\tabcolsep}{2pt} 
\renewcommand{\arraystretch}{0.9} 
\begin{tabular}{@{}l@{\hspace{20pt}}l@{}}
\toprule
\textbf{Model} & \textbf{Hyperparameters (Values)} \\
\midrule
Random Forest & \xt{n\_estimators}: [100, 200, 300, 400, 500] \\
              & \xt{max\_depth}: [None, 5, 10, 15, 20] \\
              & \xt{min\_samples\_split}: [2, 5, 10] \\
              & \xt{min\_samples\_leaf}: [1, 2, 4] \\
              & \xt{rfe\_n\_features\_to\_select}: [4,5,6,7,8,9] \\
\midrule
XGBoost       & \xt{n\_estimators}: [100, 200, 300, 400, 500] \\
              & \xt{max\_depth}: [3, 6, 10, 15, 20] \\
              & \xt{learning\_rate}: [0.001, 0.01, 0.1] \\
              & \xt{subsample}: [0.6, 0.8, 1.0] \\
              & \xt{rfe\_n\_features\_to\_select}: [4,5,6,7,8,9] \\
\midrule
K-Nearest Neighbors & \xt{n\_neighbors}: [2, 3, 5, 7] \\
                          & \xt{weights}: ['uniform', 'distance'] \\
                          & \xt{p}: [1, 2] \\
                          & \xt{selector\_k}: [4,5,6,7,8,9] \\
\midrule
Logistic Regression & \xt{C}: [0.01, 0.1, 1, 10, 50, 100] \\
                    & \xt{solver}: ['liblinear'] \\
\midrule
SVM & \xt{C}: [0.01, 0.1, 1, 3, 5, 10] \\
    & \xt{kernel}: ['linear', 'rbf', 'poly'] \\
    & \xt{gamma}: ['scale', 'auto'] \\
    & \xt{rfe\_n\_features\_to\_select}: [4,5,6,7,8,9] \\
\midrule
Gradient Boosting & \xt{n\_estimators}: [100, 200, 300, 400, 500] \\
                  & \xt{learning\_rate}: [0.001, 0.01, 0.1, 0.2] \\
                  & \xt{max\_depth}: [3, 5, 7] \\
                  & \xt{rfe\_n\_features\_to\_select}: [4,5,6,7,8,9] \\
\midrule
AdaBoost & \xt{n\_estimators}: [100, 200, 300, 400, 500] \\
         & \xt{learning\_rate}: [0.001, 0.01, 0.1, 1] \\
         & \xt{rfe\_n\_features\_to\_select}: [4,5,6,7,8,9] \\
\midrule
Extra Trees & \xt{n\_estimators}: [100, 200, 300, 400, 500] \\
            & \xt{max\_depth}: [None, 10, 20, 30] \\
            & \xt{min\_samples\_split}: [2, 5, 10] \\
            & \xt{min\_samples\_leaf}: [1, 2, 4] \\
            & \xt{rfe\_n\_features\_to\_select}: [4,5,6,7,8,9] \\
\bottomrule
\end{tabular}
\label{tab3}
\end{table}


\section{RESULTS} \label{sec:res}
In this section, we first report the results of our Topic Modeling analysis alongside visualizations, and second, we showcase the results for the prediction of the upward and downward market movements for Bitcoin and Ethereum in the test set. We compare the best models accuracies with the baselines to exhibit the potential predictive power that blockchain-embedded sentiment holds. Then, we discuss and evaluate our results in terms of usefulness and value.

\subsection{Topic Modeling} \label{5a}

For classifying text content, we use unsupervised machine learning techniques, particularly topic modeling, to uncover the hidden semantic patterns in the blockchain text corpuses of both Bitcoin and Ethereum. We use Latent Dirichlet Allocation (\textit{LDA}) \cite{blei2006}, \cite{tharwat} and \textit{BERTopic} \cite{grootendorst2022}, to enable a deeper, more contextually informed automatic classification of the texts than traditional quantitative methods which are based on the mere occurrence of counts and format types or the qualitative inspection of the embedded blockchain documents. \textit{LDA} offers a probabilistic framework for discerning broad themes, while \textit{BERTopic}, utilizes \textit{BERT} embeddings and sophisticated clustering techniques (\textit{UMAP} and \textit{HDBSCAN} \cite{mcinnes2018}), providing a more context-rich analysis.
Table \ref{tab2} presents the top 4 thematic topics of \textit{LDA} and \textit{BERTopic} analyses respectively for \textit{btc-corpus}, \textit{btc-financial-corpus}, \textit{eth-corpus} and \textit{eth-financial corpus}.

\begin{table*}[htbp]
\centering
\caption{Topic Modeling for blockchains' corpuses}
\begin{tabular}
{@{}l@{\hspace{2pt}}*4l@{}}
\toprule
\textbf{Topic} & \multicolumn{4}{c}{\textbf{LDA}} \\
\cmidrule{2-5}
 & \textbf{btc-corpus} & \textbf{btc-corpus-financial} & \textbf{eth-corpus} & \textbf{eth-corpus-financial}  \\
\midrule
1 & \makecell[tl]{peer, bitcoin, cash, \\electronic, email, title} & \makecell[tl]{function, return, length,\\ test, null, data} & \makecell[tl]{bart, grok, data,\\ sell, action, refer} & \makecell[tl]{token, revoke, approval, \\phishing, price, data} \\
\midrule
2 & \makecell[tl]{test, love, function, \\hello, world, return} & \makecell[tl]{link, long, money, \\type, title, description} & \makecell[tl]{twitter, service, charge, \\wallet, address, free} & \makecell[tl]{sent, coin, buy, \\message, scam', address} \\
\midrule
3 & \makecell[tl]{private, buy, atom, \\gaia, href, contact} & \makecell[tl]{blockchain, today, signal, \\satoshi, carving, hello} & \makecell[tl]{data, tick, mint,\\ tree, cash, identity} & \makecell[tl]{volume, ticket, type, \\false, uint, function}\\
\midrule
4 & \makecell[tl]{doge, blockchain, today, \\signal, luna, terra} & \makecell[tl]{bitcoin, wallet, recovery, \\decryption, buy, double} & \makecell[tl]{data, base, image, \\text, html, payout} & \makecell[tl]{money, funds, send,\\ address, return, help}  \\
\midrule

&\multicolumn{4}{c}{\textbf{BERTopic}}\\
\cmidrule{2-5}
 & \textbf{btc-corpus} & \textbf{btc-corpus-financial} & \textbf{eth-corpus} & \textbf{eth-corpus-financial}  \\
\midrule
1 & wallet\_decryption\_recovery\_bitcoin & saving\_contact\_asset\_backed & nonce\_json\_application\_mint & coinfirm\_live\_corporation \_officially\\
\midrule
2 & swap\_rune\_thor\_kraken & discuss\_gmail\_reward\_asset & image\_base\_data\_hose & charge\_service\_delivery\_initiative\\
\midrule
3 & peer\_electronic\_cash\_bitcoin & wallet\_decryption\_recovery\_bitcoin & nonce\_perc\_mint\_tick & online\_mining\_ethereum\_comp\\
\midrule
4 & private\_buy \_spendable\_borrowed & buy\_bitcoin\_bitcoins\_f*cker & facet\_mint\_tick\_data & free\_wallet\_religion\_leak\\
\bottomrule
\end{tabular}
\label{tab2}
\end{table*}\vspace*{-0.3cm}


The LDA analysis shows a diverse range of topics within the blockchains text corpuses. For instance, Topic 1 of \textit{btc-corpus} is heavily focused on Bitcoin, with key terms like ``\xt{peer}'', ``\xt{cash}'', and ``\xt{electronic}'', which are the foundational aspects of Bitcoin as a digital currency. Topic 2 is more aligned with programming, evident from words like ``\xt{test}'', ``\xt{function}'', and ``\xt{hello world}''. In Topic 3, there's a clear inclination towards privacy and transactional themes, indicated by terms like ``\xt{private}'' and ``\xt{buy}''. Topic 4 suggests a concentration on specific cryptocurrencies such as ``\xt{doge}'' and ``\xt{luna}'' alongside with words such as ``\xt{today}'' and ``\xt{signal}'' which indicate that the financial signals have been recognized and classified in the same topic. Contrastingly, the \textit{BERTopic} offers a more precise and focused thematic understanding. The topics are more coherent, centering around specific areas like wallet recovery and specific cryptocurrency transactions. Unlike \textit{LDA}'s broader, more varied themes, \textit{BERTopic} seems to provide a sharper and more detailed thematic analysis.\par
These kinds of observations are similar to the rest of the \textit{LDA} and \textit{BERTopic} analyses and for the ETH corpuses as well. We are more interested in examining the financial corpuses generated topics and comparing them with the ones generated for the whole corpuses, given that the former are the ones we will use for our second task. 
To that end, we indicatively plot the intertopic distance maps of \textit{btc-corpus} and \textit{btc-financial-corpus} to showcase the concentration of \textit{BERT's} topics after the filtering and preprocessing of our datasets.

\begin{figure}[tbp]
\centering
\begin{minipage}[b]{\columnwidth} 
    \centering
    \vspace{-0.38cm}
    \includegraphics[width=0.8\linewidth]{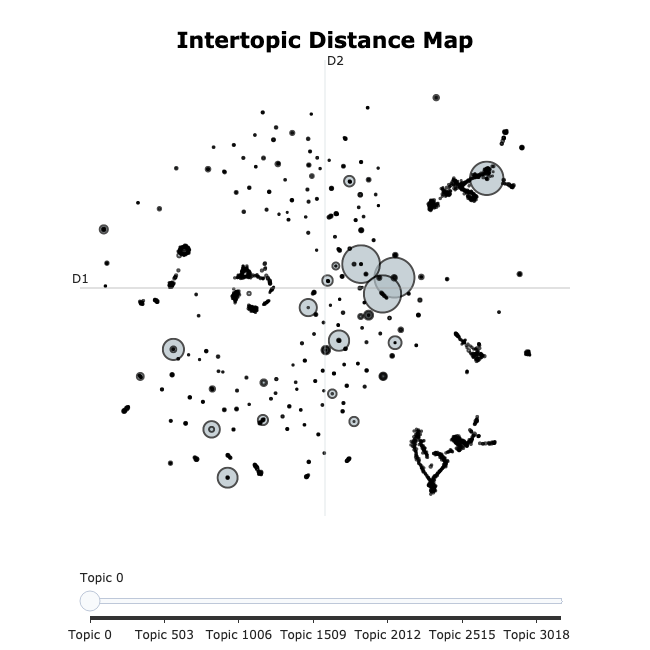}
\end{minipage}
\hfill
\begin{minipage}[b]{\columnwidth} 
    \centering
    \includegraphics[width=0.8\linewidth]{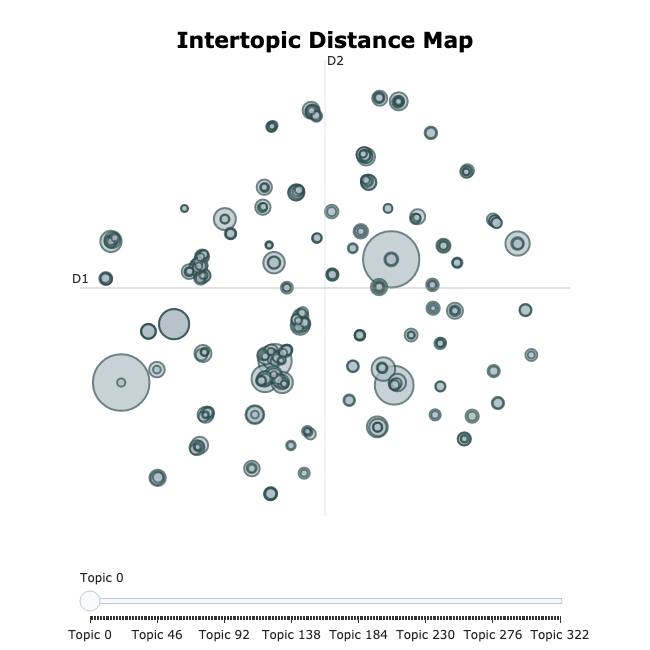}
\end{minipage}
\caption{Comparison of Bitcoin corpuses topics' clustering. }
\label{fig1}
\end{figure}

Figure \ref{fig1} illustrates this comparison. We observe that the total number of topics from 3018 is reduced to just 322 and that the topics are much more concentrated in the financial corpus. Unlike the topics in the intertopic distance map of the \textit{btc-corpus}, they are less scattered across the plot, with significantly less outliers, providing a wider range of thematic groupings. Thus, we verify the usefulness of the filtering and preprocessing steps we performed in the previous subsection.
Overall, after the topic modeling analysis, all blockchain text data have been automatically classified based on their thematic and contextual properties.

\subsection{ML Classifier and Baseline Results} \label{5b}

To evaluate our methodology, we select the models with the best cross-validation accuracy after training and hyperparameter tuning as described in subsection \ref{MLsubsec} and test them in our out-of-sample (test) set. We choose the \textit{accuracy metric} for optimization as the most intuitive and widely used classification metric in the literature. Given that our datasets are balanced, the accuracy metric can be used safely, without our results being misleading in terms of significance interpretation \cite{loginova2024}. 
The experiment configurations include the following settings; first, only using sentiment features and second, also including price features, i.e. the current day's price movement and it's current logarithmic return. The reason for that is to examine how our ML models will behave and to what extend the blockchain's sentiment is useful. For the third configuration, we use a variance of the financial corpuses presented in Section \ref{sec:method}. It considers minimal preprocessing and filtering steps, as we also want to examine if keeping information that we may consider redundant can be useful for the ML models after all. In Table \ref{tab:metrics_side_by_side}, we refer to the produced sentiment of this configuration as \textit{Raw Sentiment}.\par

We also compare our results in terms of accuracy with baseline models. However, given the nature of our dataset, we can consider various baseline models. One could argue that since we cannot know the distribution of 1s (price went up) and 0s (price went down) for the test set beforehand, the baseline is the naïve approach of $50\%$, i.e., predicting randomly if the price will go up or down. However, given that the test set may not contain an equal amount of ups and downs, i.e., it is imbalanced, we could also argue that, a more difficult to surpass, baseline accuracy for ML models is the one of continuously predicting the majority class, e.g., having a lucky guess that the price will always go up. This is necessary for a more robust evaluation, as financial markets often experience periods, where assets tend to either always move upwards or downwards \cite{maheu2000}. Thus, depending on the dataset's timespan, it can be very straightforward to obtain a high accuracy if the market is moving, e.g., $90\%$ of the time in an upwards direction, by always predicting a positive direction for all observations \cite{sun2009, loginova2024}. Hence, as mentioned in \cite{loginova2024}, given that most previous studies do not report on the class imbalance, model performance is difficult to assess across multiple studies. Therefore, we consider and evaluate our results in respect to both the aforementioned baselines alongside with a third one, which is the best ML models configurations that only take into consideration the price features. Table \ref{tab:metrics_side_by_side} presents the test set's full classification metrics results for the baselines as well as for our best ML models for the aforementioned settings for both Bitcoin and Ethereum price movements. 

\begin{table*}[tbp]
\centering
\caption{1-Day Ahead Price Movement Predictions’ Metrics for Bitcoin and Ethereum}
\setlength{\tabcolsep}{3pt} 
\renewcommand{\arraystretch}{1.1} 
\begin{minipage}[t]{0.48\textwidth}
    \centering
    \textbf{Bitcoin Results}\\
    \vspace{2pt} 
    \begin{tabular}{@{}llccc@{}}
    \toprule
    \textbf{Model} & \textbf{Accuracy} & \textbf{Precision} & \textbf{Recall} & \textbf{F1-Score} \\
    \midrule
    Random Guess              & 50.00\% & - & - & - \\
    Lucky Guess               & 52.30\% & 27\% & 52\% & 36\% \\
    XGBC+Price                & 52.96\% & 56\% & 53\% & 39\% \\
    LRC+Sentiment             & 56.25\% & 58\% & 56\% & 51\% \\
    RFC+Sentiment+Price       & 60.53\% & 60\% & 61\% & 60\% \\
    GBC+Raw Sentiment         & 53.29\% & 59\% & 53\% & 40\% \\
    RFC+Raw Sentiment+Price   & 52.96\% & 53\% & 53\% & 52\% \\
    \bottomrule
    \end{tabular}
\end{minipage}
\hfill
\begin{minipage}[t]{0.48\textwidth}
    \centering
    \textbf{Ethereum Results}\\
    \vspace{2pt} 
    \begin{tabular}{@{}llccc@{}}
    \toprule
    \textbf{Model} & \textbf{Accuracy} & \textbf{Precision} & \textbf{Recall} & \textbf{F1-Score} \\
    \midrule
    Random Guess              & 50.00\% & - & - & - \\
    Lucky Guess               & 52.96\% & 28\% & 53\% & 37\% \\
    ABC+Price                 & 51.32\% & 55\% & 51\% & 47\% \\
    XGBC+Sentiment            & 51.64\% & 51\% & 52\% & 51\% \\
    KNNC+Sentiment+Price      & 52.30\% & 53\% & 52\% & 52\% \\
    KNNC+Raw Sentiment        & 50.00\% & 51\% & 50\% & 50\% \\
    ABC+Raw Sentiment+Price   & 56.25\% & 56\% & 56\% & 55\% \\
    \bottomrule
    \end{tabular}
\end{minipage}
\label{tab:metrics_side_by_side}
\end{table*}



\subsection{Results Evaluation \& Discussion} \label{sec:discussion}

From the results presented in Subsection \ref{5b}, we observe that all classification metrics for Bitcoin predictions that utilize blockchain sentiment features achieve better results than all the baselines. This verifies our initial hypothesis that texts that are arbitrarily embedded in blockchains can indeed have predictive power on cryptocurrency price fluctuations. When combined with price data, we observe that the metrics get even more improved, as expected. Interestingly, for Ethereum our approach doesn't surpass the second baseline in three out of our four experiments that include sentiment features in terms of accuracy. Counterintuitively, the use of raw sentiment in combination with price data seems to slightly improve Ethereum results, even though it does not contribute much when used individually. However, individually weak features that may seem unhelpful on their own can significantly enhance performance when combined \cite{guyon2003}. Nonetheless, Bitcoin’s accuracy has systematically better improvement than that of Ethereum’s predictions because the aggregated texts hold sentiment more directed towards Bitcoin. This can also be seen from our analysis in Section \ref{5a}, where most of the top topics recognized by both the LDA and the BERTopic methodologies mainly involve Bitcoin in their thematology.\par
Further, the most important sentiment features selected by the best model, according to the RFE criterion, were VADER's \textit{compound, neg}, and \textit{neu} for both cryptocurrencies. For Bitcoin's model, TextBlob's \textit{subjectivity} and \textit{polarity} features were also selected. Notably, RFE's elimination of the specialized sentiment from both cryptoBERT variations, demonstrates a clear preference of the ML algorithms for the rule- and lexicon-based produced features as the most impactful for predictions. A possible explanation for this is that VADER and TextBlob produce continuous signals, enabling the ML models to capture subtle gradations of sentiment more effectively. Conversely, the specialized BERT variants did not fully adapt to the nuanced style of embedded blockchain texts, indicating the need for more fine-grained models tailored to the particular blockchain domain.

Lastly, we report improved results for Bitcoin price movement predictions in the same order of magnitude as state-of-the-art models \cite{loginova2024, passalis2022, gurrib2022} that include accuracy results. This demonstrates that blockchain embedded content offers a novel, previous unexplored, yet reliable and powerful source of harvesting text data that carries sentiment with significant (equal to costly online resources) predictive power over cryptocurrency price movements.
\section{CONCLUSION} \label{sec:conclusion}

In this paper, we used the sentiment that can be extracted from the arbitrarily embedded messages found in blockchains to predict the price movement of the two most famous cryptocurrencies, Bitcoin and Ethereum. We quantified its predictive power by feeding Machine Learning Classifiers with the text sentiments, and achieved up to $10.53\%$ improved accuracy compared to various baseline models for 1-day ahead predictions. This matches state-of-the-art prediction models that have thus far relied on proprietary data. Hidden sentiment across blockchains was found to have greater impact on Bitcoin's price movement predictions compared to Ethereum's. Furthermore, we automatically discovered hidden semantic patterns and identified topics that exist within the embedded text corpus in both the Bitcoin and Ethereum blockchains through topic modeling (unsupervised machine learning classification). In this way, we quantitatively classified every arbitrary content based on its contextual properties. Our analysis identifies significant source of freely available, permanent and immutable data as a significant predictor of cryptocurrency price movements. Regarding future research, our work lays the foundation for a deeper understanding of the inherent volatility in the field of blockchain analytics by introducing Natural Language Processing and domain-specific knowledge as a robust framework to process blockchain data for sentiment and thematic analyses. 

\bibliography{bibliography}{}

\begin{thebibliography}{10}
\providecommand{\url}[1]{#1}
\csname url@samestyle\endcsname
\providecommand{\newblock}{\relax}
\providecommand{\bibinfo}[2]{#2}
\providecommand{\BIBentrySTDinterwordspacing}{\spaceskip=0pt\relax}
\providecommand{\BIBentryALTinterwordstretchfactor}{4}
\providecommand{\BIBentryALTinterwordspacing}{\spaceskip=\fontdimen2\font plus
\BIBentryALTinterwordstretchfactor\fontdimen3\font minus
  \fontdimen4\font\relax}
\providecommand{\BIBforeignlanguage}[2]{{%
\expandafter\ifx\csname l@#1\endcsname\relax
\typeout{** WARNING: IEEEtran.bst: No hyphenation pattern has been}%
\typeout{** loaded for the language `#1'. Using the pattern for}%
\typeout{** the default language instead.}%
\else
\language=\csname l@#1\endcsname
\fi
#2}}
\providecommand{\BIBdecl}{\relax}
\BIBdecl

\bibitem{nakamoto2009}
\BIBentryALTinterwordspacing
S.~Nakamoto, ``{Bitcoin: A Peer-to-Peer Electronic Cash System},'' May 2009.
  [Online]. Available: \url{http://www.bitcoin.org/bitcoin.pdf}
\BIBentrySTDinterwordspacing

\bibitem{matzutt2016}
\BIBentryALTinterwordspacing
R.~Matzutt, O.~Hohlfeld, M.~Henze, R.~Rawiel, J.~H. Ziegeldorf, and K.~Wehrle,
  ``{POSTER: I Don't Want That Content! On the Risks of Exploiting Bitcoin's
  Blockchain as a Content Store},'' in \emph{Proceedings of the 2016 ACM SIGSAC
  Conference on Computer and Communications Security}, ser. CCS '16.\hskip 1em
  plus 0.5em minus 0.4em\relax New York, NY, USA: Association for Computing
  Machinery, 2016, p. 1769–1771. [Online]. Available:
  \url{https://doi.org/10.1145/2976749.2989059}
\BIBentrySTDinterwordspacing

\bibitem{matzutt2018}
R.~Matzutt, J.~Hiller, M.~Henze, J.~H. Ziegeldorf, D.~M{\"u}llmann,
  O.~Hohlfeld, and K.~Wehrle, ``{A Quantitative Analysis of the Impact of
  Arbitrary Blockchain Content on Bitcoin},'' in \emph{Financial Cryptography
  and Data Security}, S.~Meiklejohn and K.~Sako, Eds.\hskip 1em plus 0.5em
  minus 0.4em\relax Berlin, Heidelberg: Springer Berlin Heidelberg, 2018, pp.
  420--438.

\bibitem{Shirriff2014}
K.~Shirriff, ``{Hidden surprises in the Bitcoin blockchain and how they are
  stored: Nelson Mandela, Wikileaks, photos, and Python software},''
  \url{http://www.righto.com/2014/02/ascii-bernanke-wikileaks-photographs.html},
  2014, online; accessed 2024-11-11.

\bibitem{sward2018}
A.~Sward, I.~Vecna, and F.~Stonedahl, ``{Data Insertion in Bitcoin's
  Blockchain},'' \emph{Ledger}, vol.~3, 04 2018.

\bibitem{gregoriadis2022}
\BIBentryALTinterwordspacing
M.~Gregoriadis, R.~Muth, and M.~Florian, ``{Analysis of Arbitrary Content on
  Blockchain-Based Systems using BigQuery},'' in \emph{Companion Proceedings of
  the Web Conference 2022}, ser. WWW ’22.\hskip 1em plus 0.5em minus
  0.4em\relax ACM, Apr. 2022. [Online]. Available:
  \url{http://dx.doi.org/10.1145/3487553.3524628}
\BIBentrySTDinterwordspacing

\bibitem{bartoletti2017}
M.~Bartoletti and L.~Pompianu, ``{An Analysis of Bitcoin OP{\_}RETURN
  Metadata},'' in \emph{Financial Cryptography and Data Security}, M.~Brenner,
  K.~Rohloff, J.~Bonneau, A.~Miller, P.~Y. Ryan, V.~Teague, A.~Bracciali,
  M.~Sala, F.~Pintore, and M.~Jakobsson, Eds.\hskip 1em plus 0.5em minus
  0.4em\relax Cham: Springer International Publishing, 2017, pp. 218--230.

\bibitem{bartoletti2019}
\BIBentryALTinterwordspacing
M.~Bartoletti, B.~Bellomy, and L.~Pompianu, ``{A Journey into Bitcoin
  Metadata},'' \emph{Journal of Grid Computing}, vol.~17, no.~1, pp. 3--22,
  2019. [Online]. Available: \url{https://doi.org/10.1007/s10723-019-09473-3}
\BIBentrySTDinterwordspacing

\bibitem{scheid2023}
E.~J. Scheid, S.~Küng, M.~F. Franco, and B.~Stiller, ``{Opening Pandora's Box:
  An Analysis of the Usage of the Data Field in Blockchains},'' in \emph{2023
  Fifth International Conference on Blockchain Computing and Applications
  (BCCA)}, 2023, pp. 369--376.

\bibitem{rodwald2021}
P.~Rodwald, ``{An Analysis of Data Hidden in Bitcoin Addresses},'' in
  \emph{Theory and Engineering of Dependable Computer Systems and Networks},
  W.~Zamojski, J.~Mazurkiewicz, J.~Sugier, T.~Walkowiak, and J.~Kacprzyk,
  Eds.\hskip 1em plus 0.5em minus 0.4em\relax Cham: Springer International
  Publishing, 2021, pp. 369--379.

\bibitem{blei2006}
\BIBentryALTinterwordspacing
D.~M. Blei and J.~D. Lafferty, ``Dynamic topic models,'' in \emph{Proceedings
  of the 23rd International Conference on Machine Learning}, ser. ICML
  '06.\hskip 1em plus 0.5em minus 0.4em\relax New York, NY, USA: Association
  for Computing Machinery, 2006, p. 113–120. [Online]. Available:
  \url{https://doi.org/10.1145/1143844.1143859}
\BIBentrySTDinterwordspacing

\bibitem{tharwat}
\BIBentryALTinterwordspacing
A.~Tharwat, T.~Gaber, A.~Ibrahim, and A.~E. Hassanien, ``{Linear discriminant
  analysis: A detailed tutorial},'' \emph{AI Commun.}, vol.~30, no.~2, p.
  169–190, Jan. 2017. [Online]. Available:
  \url{https://doi.org/10.3233/AIC-170729}
\BIBentrySTDinterwordspacing

\bibitem{grootendorst2022}
\BIBentryALTinterwordspacing
M.~Grootendorst, ``{BERTopic: Neural topic modeling with a class-based TF-IDF
  procedure},'' 2022. [Online]. Available:
  \url{https://arxiv.org/abs/2203.05794}
\BIBentrySTDinterwordspacing

\bibitem{raju2020}
\BIBentryALTinterwordspacing
S.~M. Raju and A.~M. Tarif, ``{Real-Time Prediction of Bitcoin Price using
  Machine Learning Techniques and Public Sentiment Analysis},'' 2020. [Online].
  Available: \url{https://arxiv.org/abs/2006.14473}
\BIBentrySTDinterwordspacing

\bibitem{georgoula2015}
I.~Georgoula, D.~Pournarakis, C.~Bilanakos, D.~Sotiropoulos, and G.~Giaglis,
  ``\BIBforeignlanguage{Undefined}{{Using Time-Series and Sentiment Analysis to
  Detect the Determinants of Bitcoin Prices}},'' in
  \emph{\BIBforeignlanguage{Undefined}{9th Mediterranean Conference on
  Information Systems, MCIS 2015, Samos, Greece, October 2-5, 2015.
  Proceedings}}, 2015, p.~20.

\bibitem{sattarov2020}
O.~Sattarov, H.~S. Jeon, R.~Oh, and J.~D. Lee, ``{Forecasting Bitcoin Price
  Fluctuation by Twitter Sentiment Analysis},'' in \emph{2020 International
  Conference on Information Science and Communications Technologies (ICISCT)},
  2020, pp. 1--4.

\bibitem{loginova2024}
\BIBentryALTinterwordspacing
E.~Loginova, W.~K. Tsang, G.~van Heijningen, L.-P. Kerkhove, and D.~F. Benoit,
  ``Forecasting directional bitcoin price returns using aspect-based sentiment
  analysis on online text data,'' \emph{Machine Learning}, vol. 113, no.~7, pp.
  4761--4784, 2024. [Online]. Available:
  \url{https://doi.org/10.1007/s10994-021-06095-3}
\BIBentrySTDinterwordspacing

\bibitem{kilimci2020}
Z.~Kilimci, ``{Sentiment Analysis Based Direction Prediction in Bitcoin using
  Deep Learning Algorithms and Word Embedding Models},'' \emph{International
  Journal of Intelligent Systems and Applications}, vol.~8, pp. 60--65, 06
  2020.

\bibitem{kumari2023}
S.~Kumari, V.~Kumar, A.~Sharmila, C.~R. Murthy, N.~Ahlawat, and G.~Manoharan,
  ``{Blockchain Based E-Analysis of Social Media Forums for Crypto Currency
  Phase Shifts},'' in \emph{2023 5th International Conference on Inventive
  Research in Computing Applications (ICIRCA)}, 2023, pp. 1222--1225.

\bibitem{coulter2022}
K.~Coulter, ``The impact of news media on bitcoin prices: modelling data driven
  discourses in the crypto-economy with natural language processing,''
  \emph{Royal Society Open Science}, vol.~9, 04 2022.

\bibitem{Bhatt2023}
\BIBentryALTinterwordspacing
S.~Bhatt, M.~Ghazanfar, and M.~Amirhosseini, ``{Sentiment-Driven Cryptocurrency
  Price Prediction: A Machine Learning Approach Utilizing Historical Data and
  Social Media Sentiment Analysis},'' \emph{Machine Learning and Applications:
  An International Journal (MLAIJ)}, vol.~10, no. 2/3, pp. 1--15, 2023.
  [Online]. Available: \url{https://doi.org/10.5121/mlaij.2023.10301}
\BIBentrySTDinterwordspacing

\bibitem{passalis2022}
\BIBentryALTinterwordspacing
N.~Passalis, L.~Avramelou, S.~Seficha, A.~Tsantekidis, S.~Doropoulos,
  G.~Makris, and A.~Tefas, ``{Multisource Financial Sentiment Analysis for
  Detecting Bitcoin Price Change Indications Using Deep Learning},''
  \emph{Neural Computing and Applications}, vol.~34, no.~22, pp.
  19\,441--19\,452, 2022. [Online]. Available:
  \url{https://doi.org/10.1007/s00521-022-07509-6}
\BIBentrySTDinterwordspacing

\bibitem{gurrib2022}
\BIBentryALTinterwordspacing
I.~Gurrib and F.~Kamalov, ``{Predicting Bitcoin Price Movements Using Sentiment
  Analysis: A Machine Learning Approach},'' \emph{Studies in Economics and
  Finance}, vol.~39, no.~3, pp. 347--364, 2022. [Online]. Available:
  \url{https://doi.org/10.1108/SEF-07-2021-0293}
\BIBentrySTDinterwordspacing

\bibitem{hutto2014}
\BIBentryALTinterwordspacing
C.~Hutto and E.~Gilbert, ``{VADER: A Parsimonious Rule-Based Model for
  Sentiment Analysis of Social Media Text},'' \emph{Proceedings of the
  International AAAI Conference on Web and Social Media}, vol.~8, no.~1, pp.
  216--225, May 2014. [Online]. Available:
  \url{https://ojs.aaai.org/index.php/ICWSM/article/view/14550}
\BIBentrySTDinterwordspacing

\bibitem{loria2018textblob}
S.~Loria, ``{TextBlob Documentation},'' \emph{Release 0.15}, vol.~2, 2018.

\bibitem{devlin2019}
\BIBentryALTinterwordspacing
J.~Devlin, M.-W. Chang, K.~Lee, and K.~Toutanova, ``{{BERT}: Pre-training of
  Deep Bidirectional Transformers for Language Understanding},'' in
  \emph{Proceedings of the 2019 Conference of the North {A}merican Chapter of
  the Association for Computational Linguistics: Human Language Technologies,
  Volume 1 (Long and Short Papers)}, J.~Burstein, C.~Doran, and T.~Solorio,
  Eds.\hskip 1em plus 0.5em minus 0.4em\relax Minneapolis, Minnesota:
  Association for Computational Linguistics, Jun. 2019, pp. 4171--4186.
  [Online]. Available: \url{https://aclanthology.org/N19-1423}
\BIBentrySTDinterwordspacing

\bibitem{matzutt2022}
R.~Matzutt, V.~Ahlrichs, J.~Pennekamp, R.~Karwacik, and K.~Wehrle, ``{A
  Moderation Framework for the Swift and Transparent Removal of Illicit
  Blockchain Content},'' in \emph{2022 IEEE International Conference on
  Blockchain and Cryptocurrency (ICBC)}, 2022, pp. 1--9.

\bibitem{politis2021}
A.~Politis, K.~Doka, and N.~Koziris, ``{Ether Price Prediction Using Advanced
  Deep Learning Models},'' in \emph{2021 IEEE International Conference on
  Blockchain and Cryptocurrency (ICBC)}, 2021, pp. 1--3.

\bibitem{saad2018}
M.~Saad and A.~Mohaisen, ``Towards characterizing blockchain-based
  cryptocurrencies for highly-accurate predictions,'' in \emph{IEEE INFOCOM
  2018 - IEEE Conference on Computer Communications Workshops (INFOCOM
  WKSHPS)}, 2018, pp. 704--709.

\bibitem{akcora2018}
C.~G. Akcora, A.~K. Dey, Y.~R. Gel, and M.~Kantarcioglu, ``{Forecasting Bitcoin
  Price with Graph Chainlets},'' in \emph{Advances in Knowledge Discovery and
  Data Mining}, D.~Phung, V.~S. Tseng, G.~I. Webb, B.~Ho, M.~Ganji, and
  L.~Rashidi, Eds.\hskip 1em plus 0.5em minus 0.4em\relax Cham: Springer
  International Publishing, 2018, pp. 765--776.

\bibitem{li2020blockchaintransactiongraphbased}
\BIBentryALTinterwordspacing
X.~Li and W.~Wu, ``{A Blockchain Transaction Graph based Machine Learning
  Method for Bitcoin Price Prediction},'' 2020. [Online]. Available:
  \url{https://arxiv.org/abs/2008.09667}
\BIBentrySTDinterwordspacing

\bibitem{kleitsikas}
C.~Kleitsikas, K.~Doka, A.~Politis, and N.~Koziris, ``{Graph-Centric Crypto
  Price Prediction},'' in \emph{2023 IEEE International Conference on
  Blockchain and Cryptocurrency (ICBC)}, 2023, pp. 1--5.

\bibitem{Kok22}
C.~Koki, S.~Leonardos, and G.~Piliouras, ``{Exploring the predictability of
  cryptocurrencies via Bayesian hidden Markov models},'' \emph{Research in
  International Business and Finance}, vol.~59, p. 101554, 2022.

\bibitem{guyon2002}
\BIBentryALTinterwordspacing
I.~Guyon, J.~Weston, S.~Barnhill, and V.~Vapnik, ``{Gene Selection for Cancer
  Classification using Support Vector Machines},'' \emph{Machine Learning},
  vol.~46, no.~1, pp. 389--422, 2002. [Online]. Available:
  \url{https://doi.org/10.1023/A:1012487302797}
\BIBentrySTDinterwordspacing

\bibitem{mcinnes2018}
L.~McInnes, J.~Healy, N.~Saul, and L.~Grossberger, ``{UMAP: Uniform Manifold
  Approximation and Projection},'' \emph{Journal of Open Source Software},
  vol.~3, p. 861, 09 2018.

\bibitem{maheu2000}
\BIBentryALTinterwordspacing
J.~M. Maheu and T.~H. McCurdy, ``{Identifying Bull and Bear Markets in Stock
  Returns},'' \emph{Journal of Business \& Economic Statistics}, vol.~18,
  no.~1, pp. 100--112, 2000. [Online]. Available:
  \url{http://www.jstor.org/stable/1392140}
\BIBentrySTDinterwordspacing

\bibitem{sun2009}
\BIBentryALTinterwordspacing
Y.~SUN, A.~K.~C. WONG, and M.~S. KAMEL, ``{Classification OF Imbalanced Data: A
  Review},'' \emph{International Journal of Pattern Recognition and Artificial
  Intelligence}, vol.~23, no.~04, pp. 687--719, 2009. [Online]. Available:
  \url{https://doi.org/10.1142/S0218001409007326}
\BIBentrySTDinterwordspacing

\bibitem{guyon2003}
I.~Guyon and A.~Elisseeff, ``An introduction to variable and feature
  selection,'' \emph{J. Mach. Learn. Res.}, vol.~3, no. null, p. 1157–1182,
  Mar. 2003.

\end{thebibliography}
\bibliographystyle{IEEEtran}
\end{document}